# Classifying States of Cooking Objects Using Convolutional Neural Network

Qi Zheng

*Abstract* — Automated cooking machine is a goal for the future. The main aim is to make the cooking process easier, safer, and create human welfare. To allow robots to accurately perform the cooking activities, it is important for them to understand the cooking environment and recognize the objects, especially correctly identifying the state of the cooking objects. This will significantly improve the correctness of the following cooking recipes. In this project, several parts of the experiment were conducted to design a robust deep convolutional neural network for classifying the state of the cooking objects from scratch. The model is evaluated by using various techniques, such as adjusting architecture layers, tuning key hyperparameters, and using different optimization techniques to maximize the accuracy of state classification.

## I. INTRODUCTION

As artificial intelligence becomes more and more popular in modern society, many researchers and companies diving into this field. Automation machine has become a significant role in many companies and industries, slowly replacing human workers. This is not only in the workplace, but also home environment. Robot devices are now widely used in households that helps to raise efficiencies of daily life, such as smart home assistant and automatic robot vacuum. Kitchen robots are still rarely seen in the household or even big chain restaurant. Kitchen robots in big chain restaurant can play a significant role, where it can solve the problem of hygiene, safety, and labor.

The future aim is to invent and enhance a powerful kitchen robot capable of performing complex tasks, like cooking, by improving their learning and activities process. For the robot to execute cooking actions to achieve human like accuracy, much work has been done. Understanding grouping manipulation of motion from robotics perspective [1] and performing appropriate grasp for an object [2] allows the robot to execute more accurate and smoother actions. Pouring is a motion that often happens during the cooking process; thus, accuracy in controlling the input is essential for the flavor and result of the dish [3][4]. A proper recipe for the robot to understand and follow can ensure that the task is accomplished in the proper way [5].

Object state recognition is another challenge for the kitchen robot, as robots need to properly identity what the objects are in order to proceed. Several experiments have been done on objects' state detection [6][7][8][9][10]. However, all these proposed models are done by modifying and fine-tuning a pretrained model. The goal for this project is to exploit deep learning approach for image classification on the state of the cooking objects by designing a deep Convolutional Neutral Network from scratch. The proposed model is trained and validated on the Cooking State Recognition Challenge Dataset [11], achieving an accuracy of 66.95% on the validation dataset.

## II. DATA AND PREPROCESSING

### A. Dataset and Implementation

In the experiment, the Cooking State Recognition Challenge dataset version 2.0 was used [11] along with the additional images annotated by the students and verified by the TA. The resulting dataset contains total of 7213 (82%) images for training and 1543 (18%) images for validation. The images are three-channel RGB color images. The dataset includes a total of 18 object types (chicken/turkey, beef/pork, tomato, onion, bread, pepper, cheese, strawberry, etc.) with 11 different states (creamy paste, diced, floured, grated, juiced, julienne, mixed, other, peeled, sliced, and whole). The dataset in each training and validation consists of balanced class distribution as illustrated in Fig. 1. Balanced classes dataset is essential for classification model because imbalanced classes will cause the model to ignore the minority class in favor of the majority class, which can result in poor performance on balanced test datasets. The implementation for the proposed model was done in Python and PyTorch.

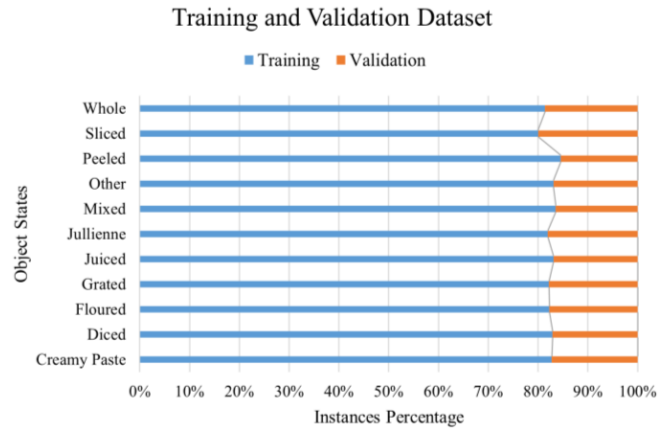

Fig. 1: Datasets Percentage Split

### B. Data Preprocessing

Data preprocessing is a critical phrase to create a robust model. It does not only allow the removal of the noise from the images and standardize the data but also increases and enhances the existing dataset to allows neural networks to learn more useful representation of features directly from the data. Especially in our case, where the training dataset is relatively small, it can easily suffer from overfitting, which can result in a very good performance on training dataset, but extremely poor performance on the dataset unseen before.

Before feeding the input images to the model, both training and validation dataset are resized to 256, center crop to 224, and then normalize using the mean and standard deviation to keep all values within a scale. Additional argumentations are applied to the training dataset, such as random horizontal and vertical flip, color jitter, random affine and gaussian blur. Detailed argumentation factors are shown in the Table I. Fig. 2 illustrated some examples of augmented images.

Table I: Overview of Data Argumentation

| Parameters | Argumentation Factor |
| --- | --- |
| Random Horizontal Flip | Probability = 0.7 |
| Random Vertical Flip | Probability = 0.3 |
| Color Jitter | Brightness = 0.2<br>Contrast = 0.2<br>Saturation = 0.2<br>Hue = 0.1 |
| Random Affine | Rotation Degree = (-20, 20)<br>Translation = (0.1, 0.1)<br>Shear = (-10, 10) |
| Gaussian Blur | 13 |

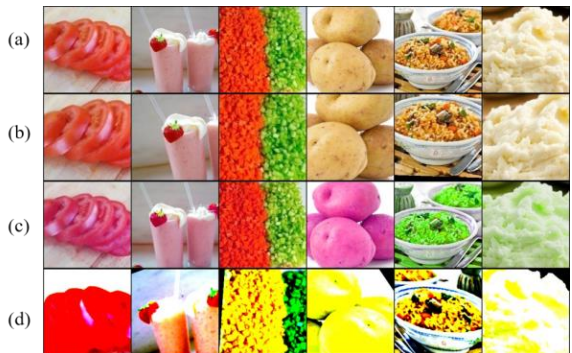

Fig. 2: Data Argumentation: (a) Original, (b) Random Affine, (c) Color Jitter, (d) Mix of data argumentation

Reflection padding was used during the experiment; however, it did not achieve a good result because reflection can significantly change the state of an object. For instance, after applied the reflection padding, a half potato with a label of "other" will be misclassified as a "whole" potato. Therefore, simple image transformation will allow model to learn better characteristics, whereas over or improper transformation will have negative impact on the performance.

## III. METHODOLOGY

The proposed model is designed and trained from scratch. Different self-organized CNN model architectures were conducted.

### A. Experiment on VGG16 Architecture

To begin with, VGG16 architecture was experimented with and analyzed. VGG16 is one of the most popular convolutional neural networks proposed by K. Simonyan and A. Zisserman [12]. It is famous for using multiple 3x3 kernel-size filters one after another with a total of 13 convolutional layers, 5 pooling layers and 3 dense layers. The model has a total of 138 million of trainable parameters, which results in a slower training process than other well-known CNN. Especially in our case, VGG16 architecture is too complex for our small data due to the large number of trainable parameters and convolutional layers, making it less efficient due to model size and training time.

### B. Proposed Model

The proposed model is adjusted using the VGG16 architecture by reducing the layers and adding additional optimization techniques. The proposed model consists of six convolutional layers follow by one 5x5 adaptive average pooling and one fully connected output layer. The model has a total of 290,283 parameters and they are all trainable parameters. The network of the proposed model is illustrated in Fig. 3.

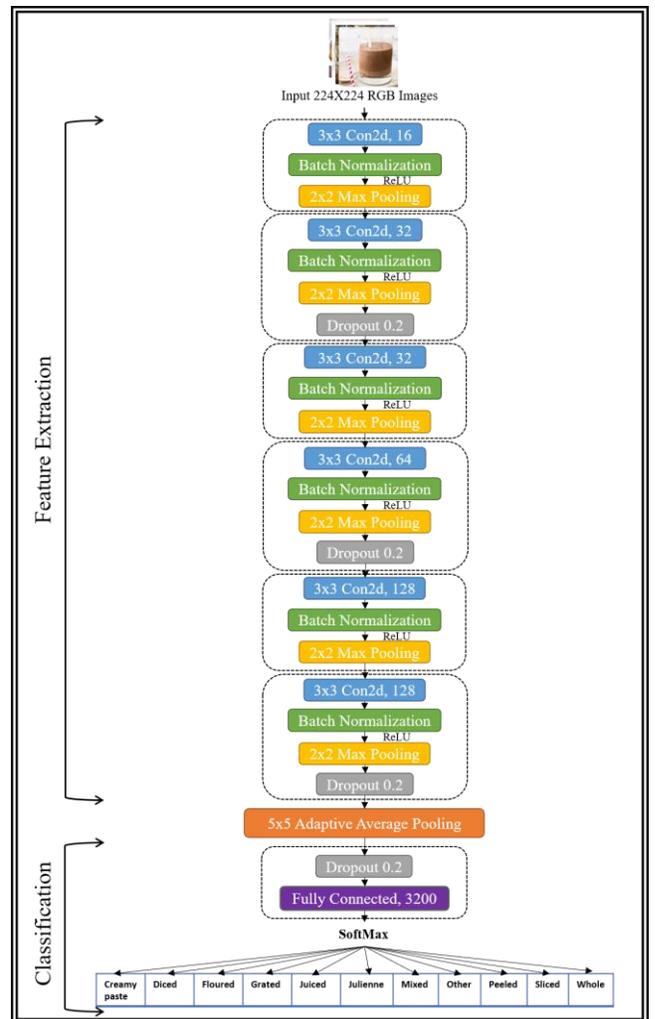

Fig. 3: Proposed Architecture

The extraction feature of the model is responsible for reducing the number of resources needed for learning process, building only the valuable information from the raw input data. This part of the model is formed by six convolutional layers. Each of the convolutional layers include a batch normalization, a Rectified Linear Unit (ReLU) as the activation function, as well as a 2x2 max pooling.

To keep the computational cost minimal, filter size 3 x 3 was applied to all the convolutional layers. Moreover, from the idea of VGG, 3x3 filter size is the most optimal choice if the model has lots of filters, which is the case for the proposed model. Initial convolutional layer has 16 filters and each of the following convolutional layer is increase by power of 2 until it reaches to 128. To avoid the model ending up with a large number of trainable parameters but still maximize the network to learn the image features, two additional convolutional layers with a size of 32 and size of 128 is added. Batch normalization is used to standardizes the inputs to a layer for each mini batch, which allows for a more predictive, stable behavior of the gradience and faster training process. The purpose of max pooling is to calculate the maximum value in each patch of the feature map, extracting the important features within the patch. To allow the model to detect more abstract concepts, smaller filter size 2x2 is selected for max pooling. The next layer is dropout layer with the value 0.2, and it is added to every other convolutional layer, allowing the model to randomly remove some nodes during the training and reduce the risk of overfitting.

Following the convolutional layers is a 5x5 adaptive average pooling, which down samples the entire feature map to a single value, helping the model summarize the presence of a feature in an image.

The proposed architecture ends with a classification section, using the result from the convolution and pooling process to classify the images into the class. The classification portion of the proposed model only consist of one fully connected layer, so a small value of 0.2 dropout layer is applied before the fully connected layer. The result of several investigations concluded that a large number of trainable parameters does not yield to a satisfactory performance due to the small dataset. Finally, the last layer of the network is a softmax activation function that output the range of probabilities between 0 and 1, commonly used in categorial probabilistic distribution.

## IV. EVALUATION AND RESULTS

The experiment mainly focused on trying different architectures and optimization techniques as well as determining the most optimal hyperparameters for the proposed model during the training. To prevent over complicated analysis, some settings are fixed. SGD optimizer is used for most the experiment, except *Effect of Different Optimizers* section, because it is commonly used for CNN. Cross-Entropy loss function is used to adjust the classification model weights during training and the aim is to minimize the loss value. Softmax is used in the output layer as the activation function to predict the multinomial probability distribution. In our case, there are total of 11 different cooking states, hence, validation loss should be less than $-ln(1/11) = 2.197$ to verify our initial model.

The default epoch size is set to 150 for all the trials throughout the experiment. Since the goal for this project is to maximize the accuracy, early stopping is observing the validation accuracy with a patience of 20 epochs during the training. High patience value is selected because of the use of learning rate decay, and it is possible that accuracy continuously increase while the learning rate decays. Throughout the training, model with maximum validation accuracy was selected.

### A. Effect of Adaptive Average Pooling

Originally, three fully connected layers was added, however, this resulted the model to be too complicated for our dataset, with only an accuracy of 58.52%. In [14], the authors concluded that replacing the fully connected layer with global average pooling lets the feature maps be easily interpreted as categories confidence maps and serve as a strategy to avoid overfitting. By experimenting the adaptive average pooling with different filter sizes, it resulted that filter size 3x3 yield a 66.04% validation accuracy and filter size 7x7 yield a 63.12% validation accuracy, whereas filter size 5x5 was able to achieve the highest accuracy of 66.49%. This part of experiment proved that replacing the two additional fully connected layer with adaptive average pooling allows the model to behave better, with significant improvement in overall accuracy.

### B. Effect of Batch Normalization

Batch normalization is a technique that can be done to standardizes the inputs to a layer for mini batch. It allows a faster and more stabilized learning process, which improves the regularization and provides faster convergence. The proposed solution in Fig. 3 without batch normalization and without dropout in convolutional layers achieved a 59.47% validation accuracy, whereas adding the batch normalization after each convolutional layer increases the validation accuracy to 66.75%. This demonstrates that batch normalization results in better regularization and yield to higher accuracy.

### C. Effect of Batch Size

Batch size controls the accuracy of the estimated error gradient when training the neural networks and is an important hyperparameter that influences the dynamics of the learning algorithm. In the experiment, four different batch size, 16, 32, 64, and 128, for SGD optimizer were conducted on epoch size 80. The comparative analysis can be found on Table II. We can observe that SGD with batch size 32 works the best for the proposed model using this dataset. Lowering the batch size to 16 decreases the accuracy by approximately 7%, whereas increasing the batch size shows a trend of decrease in accuracy and increase in loss.

Table II: Effect of Different Batch Sizes

| Batch Size | Validation Loss | Validation Accuracy |
|---|---|---|
| 16 | 1.2264 | 0.5914 |
| 32 | 1.0207 | 0.6598 |
| 64 | 1.1159 | 0.6316 |
| 128 | 1.1401 | 0.6245 |

### D. Effect of Learning Rate

Learning rate is a key configurable hyperparameter that controls the adapted speed of a model to the problem. Choosing a proper leaning rate can hugely impact the model performance. Some parts of the experiment are done on SGD optimizer with different constant learning rates and epoch-based learning rate, which is presented in Fig. 4. Different learning rate shows different graph trends between training and validation. Graph (a) shows an extreme case that when constant learning rate is too high, with value $10^{-2}$, training loss starts high but decreases substantially after first epoch and stay unchanged for rest of the epoch, which is overshooting the loss minimum. For learning rate value $10^{-3}$ in (b) and $10^{-4}$ in (c), both illustrates that convergence happened in short amount of epoch, around 30 epochs for learning rate $10^{-3}$ and around 45 epochs for learning rate $10^{-4}$. With too small of a constant learning rate, $10^{-5}$ in (d) and $10^{-6}$ in (e), convergence does not happen even after 80 epochs. Especially with learning rate $10^{-6}$, both validation loss and accuracy curves are a lot of smoother compared to the others, and training and validation cap are much wider. However, it results in a much slower convergence. Depends on the optimizer, convergence might not be found at all. Therefore, to avoid too fast or too slow convergence, a new proposed epoch-based learning rate decay was used as listed in the Table III and the corresponding performance shown in (f). Learning rate decay allows the model to start with a relatively high learning, but still benefit from the smaller learning rate to get closer to the loss minimum and learn more complex patterns.

By observing the performance of different learning rate, we can conclude that learning rate decay allows the model to reach its optimal solution within a reasonable time. Although learning rate decay in (f) shows that the model is overfitting, we conclude the architecture is complicated enough for the dataset. The problem of overfitting can be resolved by adding additional dropout, which is presented in next section *Effect of Dropout*.

TABLE III: Learning Rate Decay Based on Epoch

| Epoch | Learning Rate |
| --- | --- |
| <55 | $10^{-3}$ |
| 55 to 70 | $10^{-4}$ |
| 71 to 80 | $10^{-5}$ |
| 81 to 85 | $10^{-6}$ |
| 85 to 90 | $10^{-7}$ |
| >90 | $10^{-8}$ |

*E. Effect of Dropout*

Dropout is a technique that prevent neural networks from overfitting, where it randomly ignores some percentage of neurons to avoid the model "memorizing" interdependent set of features' weights during training. Since the proposed model only contains one fully connected layer, a low dropout rate 0.2 is set to that layer. As we can see in Fig. 5 (a), the model is still overfitting around 65 epochs. To reduce the chance of overfitting but still allow fully connected layers to successfully perform its classification task, three dropouts with value 0.2 was added to every other convolutional layer. In many of existing literature, dropout is rarely used after the convolutional layers due to small number of parameters used, and so less regularization is need. However, many researchers found that adding the dropout on the convolutional layers is not trivial, contrary, it will dramatically reduce the possibility of overfitting [14][15]. In Fig. 5 (b), we can observe such phenomenon. The huge gap between training and validation is significantly reduced and validation accuracy is slightly increased from 66.75% to 66.95%. Although it is not a tremendous improvement, the overfitting problem is mitigated.

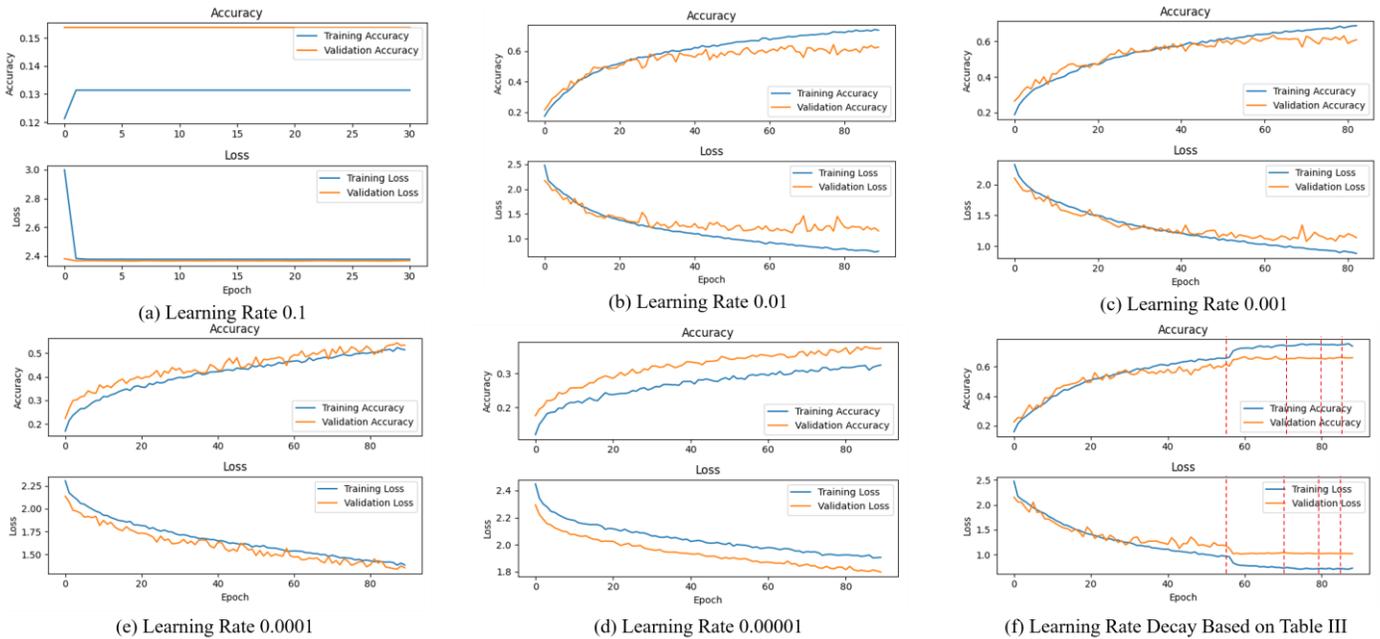

Fig. 4: Performance on Different Learning Rate

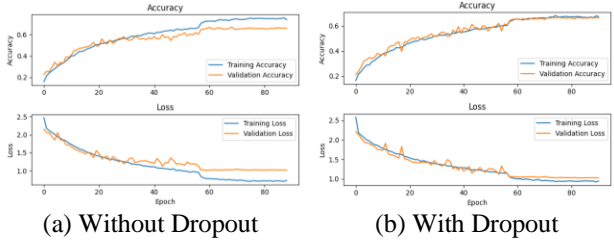

(a) Without Dropout      (b) With Dropout

Fig. 5: Training Result Dropout on Convolutional Layers

*F. Effect of Different Optimizer*

Optimizer is another essential algorithm that can significantly influence the training of the neural network. It is mainly responsible for updating the various parameters of the model such as learning rate and weights, to reduce the loss with much less effort. In this part of the experiment, eight different optimizers were tried by using the same epoch-based learning rate decay introduced in Table III. The comparative parameters of optimizers setup with its corresponding validation loss and accuracy are shown in Table IV and the details of the progress during the validation periods can be found in the Fig. 6. We can observe that the optimizer with the highest validation accuracy, 66.95%, is SGD and the optimizer with the lowest validation loss, 0.9993, is ADSG. Since higher accuracy is the goal for this project, SGD is concluded as best optimizer for the proposed model. Of course, each optimizer has its own strength and impact, by using different learning rate and different number of epochs, other optimizers may produce better accuracy.

Some interesting patterns to note. From Fig. 6, we can observe that eight difference optimizers are separated into two groups. Group one contains SGD, Adamax, ASGD, and Adagrad, showing a faster increase in validation accuracy and faster decreasing in loss, whereas group two includes Adam, AdamW, RMSprop and Adadelta, showing a slower improvement than group one. Additionally, by using the learning rate decay schedule, once learning rate decays, indicated by vertical red dish lines, there is a sharp increase in accuracy for all the optimizers, except Adadelta. It can conclude that Adadelta did not get much advantage from the smaller learning rate, on contrary, the curve is even flatter after decay.

*G. Confusion Matrix and Classification Report*

Finalized model achieved an overall 66.95% accuracy son the validation dataset. Normalized confusion matrix (%) is generated and the classification report details (precision, recall, F-1 score and x) are shown in the Fig. 6 and Table V.

It can be observed that majority of the instances can be classified into the correct categories. Highest accuracy of 82% is achieved by the states "Juiced" and "Mixed", because "juiced" have a more simple and consistent state, and "mixed" is usually colorful and therefore easier to distinguish than other states. On the other hand, "other" is found to have the lowest accuracy (31%). By observing the contents in the "other" dataset, it is easy to tell why this class has such low accuracy. State "other" contains too many non-structured objects, with some objects also being able to be classified into the many listed states and others even hard for humans to judge as illustrated in Fig. 7. All other classes have an accuracy fall in the range of 60% to 70%, which is close to the overall accuracy. If we observe the normalized confusion matrix, we can notice that "peeled" is often misclassified as "whole", because some of them still have the same overall structure. For example, a peeled orange still has the round shape as the unpeeled orange.

TABLE IV: Result of Different Optimizers

| Optimizer | Parameters Setup | Validation Loss | Validation Accuracy |
|---|---|---|---|
| SGD | momentum = 0.9 | 1.0231 | 0.6695 |
| ASGD | lambd=0.0001, alpha=0.75, t0=1000000.0, weight_decay=0 | 0.9993 | 0.6649 |
| Adadelta | rho=0.9, eps=1e-06, weight_decay=0 | 1.4979 | 0.4880 |
| Adagrad | lr_decay=0, weight_decay=0, initial_accumulator_value=0, eps=1e-10 | 1.06928 | 0.6332 |
| Adam | betas=(0.9, 0.999), eps=1e-8 | 1.2414 | 0.5768 |
| Adamax | betas=(0.9, 0.999), eps=1e-08, weight_decay=0 | 1.0661 | 0.6591 |
| AdamW | betas=(0.9, 0.999), eps=1e-08, weight_decay=0.01, amsgrad=False | 1.1908 | 0.5930 |
| RMSprop | alpha=0.99, eps=1e-08, weight_decay=0, momentum=0, centered=False | 1.2855 | 0.5605 |

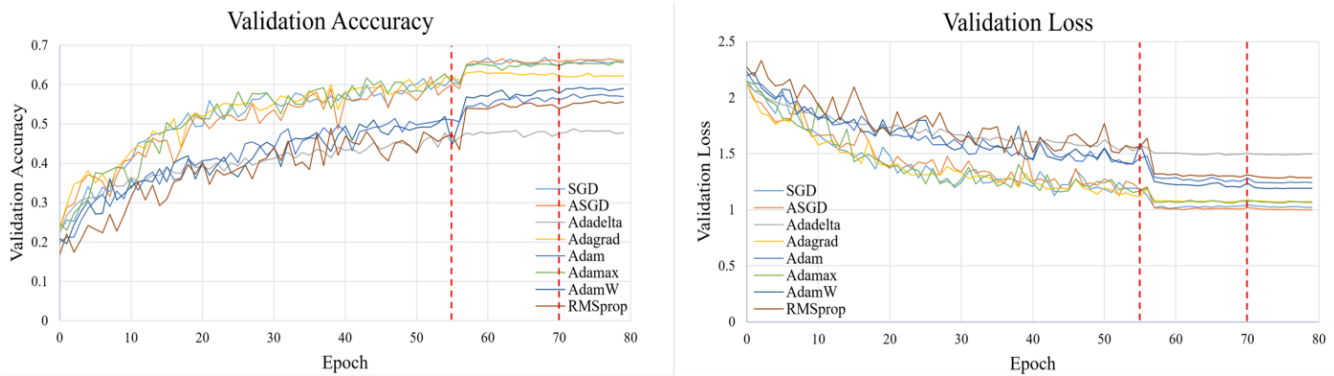

Fig. 6. Performance on Different Optimizers

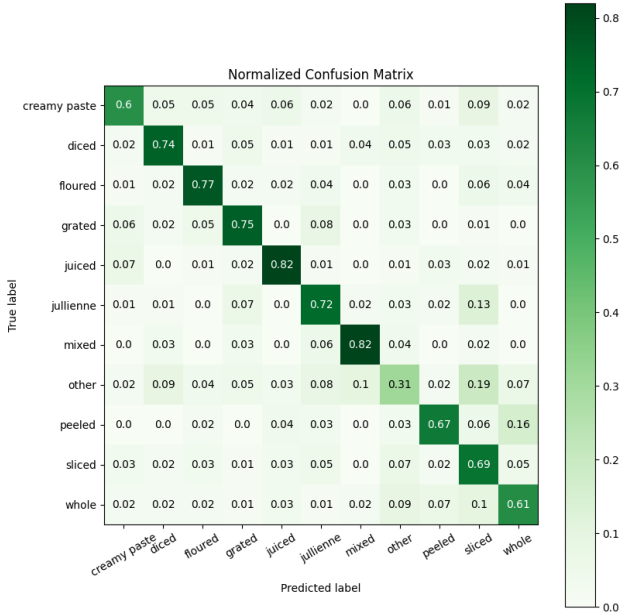

Fig. 6: Normalized Confusion Matrix on Validation Dataset

TABLE V: Classification Report on Validation Dataset

| Class | Precision | Recall | F1-score | Support |
|---|---|---|---|---|
| Creamy paste | 0.66 | 0.60 | 0.63 | 124 |
| Diced | 0.74 | 0.74 | 0.74 | 144 |
| Floured | 0.71 | 0.77 | 0.74 | 110 |
| Grated | 0.71 | 0.75 | 0.73 | 131 |
| Juiced | 0.78 | 0.82 | 0.80 | 147 |
| Julienne | 0.59 | 0.72 | 0.65 | 110 |
| Mixed | 0.75 | 0.82 | 0.78 | 99 |
| Other | 0.44 | 0.31 | 0.36 | 164 |
| Peeled | 0.67 | 0.67 | 0.67 | 102 |
| Sliced | 0.63 | 0.69 | 0.66 | 237 |
| Whole | 0.68 | 0.61 | 0.64 | 175 |
| **Average** | 0.67 | 0.68 | 0.67 | 1543 |

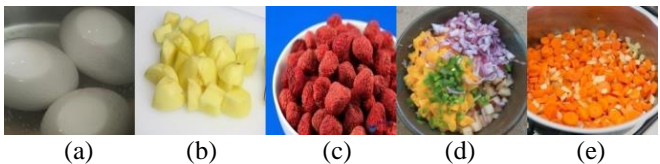

(a)     (b)     (c)     (d)     (e)

Fig. 7: Label is other, misclassified as: (a) whole, (b) diced, (c) whole, (d) mixed, (e) diced

## V. CONCLUSION

In this project, various Convolutional Neural architectures have been investigated and different optimization techniques experimented for object state classification. Experimentally, it is found that the proposed model can classify different cooking states with an overall accuracy of 66.95% on the validation set of Cooking State Recognition Challenge dataset.

Several limitations in the experiment influenced the potential of the model to achieve a higher accuracy. Upon observing some object states in dataset, it can be concluded that some data is difficult for even the human eye to differentiate and images under label "other" does not have a structured state compared to others, which resulted in largely misclassification. Although 7213 instances are given for the training, it still a small dataset. By increasing the training dataset, it allows the model to see and learn more state variations directly from the images. Additionally, since dataset is relatively small, transfer learning or fine tune a pretrain would help with learning universal representations. Lastly, training from scratch requires more iterations and time before it can reach a sufficient converge.

In the future, more experiments can be done by enhancing the image dataset, trying different filter size on the convolutional layers, and investigating deeply into the power of snapshot ensembles to improve the overall model performance on states of objects recognition.